\documentclass[runningheads]{llncs}
\RequirePackage{silence}  
\usepackage{graphicx}
\usepackage{comment}
\usepackage{amsmath,amssymb}
\usepackage{color}
\usepackage{url}
\usepackage{hyperref}
\usepackage{makecell}
\usepackage{csquotes}
\usepackage{url}
\usepackage{cleveref}
\usepackage{siunitx}
\usepackage{pifont}
\newcommand{\cmark}{\ding{51}} % Tick symbol
\newcommand{\xmark}{\ding{55}} % Cross symbol

\newif\ifreview
\reviewfalse

\ifreview
	\usepackage{lineno}

	\linenumbers
\fi

\begin{document}

%%%%%%%%%%%%%%%%%%%%% Add submission id, track, and title. %%%%%%%%%%%%%%%%%%%%%

\def\SubNumber{93}

% TODO: Please uncomment the track this paper will be submitted to, comment all other lines
%\def\GCPRTrack{Main Track}
%\def\GCPRTrack{Special Track: Pattern recognition in the life and natural sciences}
%\def\GCPRTrack{Special Track: Photogrammetry and remote sensing}
%\def\GCPRTrack{Special Track: Computer vision systems and applications}
%\def\GCPRTrack{Young Researcher's Forum}
\def\GCPRTrack{Fast Review Track}

\title{Cytoarchitecture in Words: Weakly Supervised Vision–Language Modeling for Human Brain Microscopy\thanks{
This project received funding from the European Union’s Horizon 2020 Research and Innovation Programme, grant agreement 101147319 (EBRAINS 2.0 Project), the Helmholtz Association port-folio theme “Supercomputing and Modeling for the Human Brain”, the Helmholtz Association’s Initiative and Networking Fund through the Helmholtz International BigBrain
Analytics and Learning Laboratory (HIBALL) under the Helmholtz International Lab grant agreement InterLabs-0015, from HELMHOLTZ IMAGING, a platform of the Helmholtz Information \& Data Science Incubator [X-BRAIN, grant number: ZT-I-PF-4-061], and from the
Deutsche Forschungsgemeinschaft (DFG, German Research Foundation) under the National Research Data Infrastructure – NFDI 46/1 – 501864659. Computing time was granted through JARA on the supercomputer JURECA-DC at Jülich Supercomputing Centre (JSC).
}}
% You can use \thanks for acknowledgment. Do not add any acknowledgment to the draft 
% version that is used for the review process.  
%\title{Title\thanks{XXX}}

\ifreview
	% ANONYMOUS SUBMISSION FOR REVIEW
	% DO NOT MODIFY these for the draft version that is used for the review process.
	\titlerunning{GCPR 2026 Submission \SubNumber{}. CONFIDENTIAL REVIEW COPY.}
	\authorrunning{GCPR 2026 Submission \SubNumber{}. CONFIDENTIAL REVIEW COPY.}
	\author{GCPR 2026 - \GCPRTrack{}}
	\institute{Paper ID \SubNumber}
\else
	\titlerunning{Cytoarchitecture in Words}

\author{Matthew Sutton\inst{1,2}\orcidID{0009-0001-7700-3037} \and
Katrin Amunts\inst{1,3}\orcidID{0000-0001-5828-0867} \and
Timo Dickscheid\inst{1,2,4}\orcidID{0000-0002-9051-3701} \and
Christian Schiffer\inst{1,2}\orcidID{0000-0002-2544-843X}}
\authorrunning{M. Sutton et al.}
\institute{Institute of Neuroscience and Medicine (INM-1), Research Centre Jülich, Jülich, Germany \and
Helmholtz AI, Research Centre Jülich, Jülich, Germany \and
Cécile \& Oskar Vogt Institute for Brain Research, University Hospital Düsseldorf, Düsseldorf, Germany \and
Computer Vision, Institute for Computational Visualistics, University of Koblenz, Koblenz, Germany}
\fi

\maketitle              % typeset the header of the contribution

\begin{abstract}
Vision foundation models increasingly support interactive scientific workflows, but natural-language interaction requires coupling visual representations to language. Curated image--text pairs for this coupling are scarce in many biomedical domains.
This is the case for cell-body-stained histological sections of the human brain, where microscopic image patches encode cytoarchitecture: cellular density, morphology, laminar structure, and areal organization.
We propose \emph{retrieve-and-enrich} supervision, a weakly supervised scheme for training image-conditioned language models without curated image–text pairs.
The method \emph{retrieves} label-level text from the literature via a shared anatomical label, then \emph{enriches} it with image-specific properties such as cortical layer thickness and cell density.
Labels provide training targets only, not model inputs.
We use this scheme to couple CytoNet, a cytoarchitectonic vision foundation model, to an open-weight large language model via a lightweight Flamingo-style adapter. Across 57 brain areas, the resulting model produces plausible cytoarchitectonic descriptions, supports open-set use by rejecting out-of-scope areas, and predicts the correct area for in-scope patches with 90.6\% accuracy. Removing explicit area names from generated text still leaves descriptions sufficient for a language model to recover the area in an 8-way test with 68.6\% accuracy.
A second instruction-tuned model trained with image-specific targets recovers cortical layer thickness and cell-densities from individual patches, allowing the model to read out information beyond canonical area descriptions.
Our results show that retrieve-and-enrich supervision offers a practical route to vision–language training in specialized imaging domains where labels and expert text exist but image-level captions do not.

\end{abstract}

\section{Introduction}

We propose a method for generating synthetic data for the training of a vision--language model in a domain where no paired image--text currently exists.
We demonstrate this approach by coupling a vision foundation model to language through weak, \emph{retrieve-and-enrich} supervision, training it on the derived dataset in the challenging domain of human brain cytoarchitecture.

We introduce a retrieve-and-enrich supervision strategy: each patch is first assigned a categorical label from an existing reference taxonomy (e.g. an atlas or classification scheme relevant to the domain). Crucially, these labels are used only to construct training supervision, never as inputs to the final model. For each category, canonical descriptive statements are mined and distilled from the relevant domain literature, so that the label acts as a bridge for retrieving pre-existing textual knowledge and pairing it with corresponding patches — the \emph{retrieve} step.

Because literature-derived descriptions only capture generic, category-level knowledge, supervision is further \emph{enriched} with patch-specific quantitative measurements (structural or statistical properties computed directly from the sample), allowing generated text to reflect individual instance characteristics rather than only canonical class descriptions. An LLM is then conditioned on a vision foundation model's embeddings via a cross-modal interface, producing a model capable of generating descriptive text about the canonical and instance-specific properties of a given patch, as well as answering targeted questions about it. This approach offers a general recipe for bootstrapping vision-language text generation capabilities on top of an existing domain-specific foundation model when paired image-text data is unavailable, but a reference taxonomy, a literature corpus, and computable instance-level features are available.

The human cerebral cortex is subdivided into cytoarchitectonic areas:
regions with a distinct microscopic architecture in terms of cellular density, cell morphology, and laminar (layered) organization~\cite{Amunts2015}.
Cell-body--stained histological sections acquired at microscopic resolution provide direct access to these microarchitectural signatures and underpin cortical atlases~\cite{Amunts2020}.
High-throughput histology and whole-brain reconstruction now produce terabyte-scale datasets, creating demand for scalable computational analysis~\cite{Amunts2013,Amunts2024,schiffer2025cytonet}.
CytoNet~\cite{schiffer2025cytonet}, a foundation model for human cytoarchitecture, addresses this by embedding local histological patterns into a representation space that supports downstream analysis tasks.
CytoNet learns these embeddings through contrastive training on image patches extracted from whole-slide histological human brain sections, using the spatial co-localization of image patches in a common reference coordinate system for self-supervision.
However, as is typical for such models, its outputs are exposed as high-dimensional embeddings, which are difficult to integrate into day-to-day scientific workflows, which benefit from human-readable outputs.

We instantiate this framework for human brain, generating text describing cytoarchitectonic properties of microscopic image patches.
As the reference taxonomy, we use the Julich-Brain atlas~\cite{Amunts2020}, a cytoarchitectonic reference map of the human brain in which each cortical area is delineated on histological sections and represented probabilistically to reflect its variability across individuals; area labels are automatically predicted by a CytoNet-based area classifier~\cite{schiffer2025cytonet}.
For each area, we mine the neuroanatomical literature and distill short statements describing canonical cytoarchitectonic attributes.
Patch-specific enrichment is provided by measurements including cortical layer thickness and cell density.
Conditioning an LLM on CytoNet embeddings using a Flamingo-style cross-modal interface~\cite{alayrac2022flamingo} yields a model that is able to generate text describing the canonical features of a given image patch and answer specific questions about it.

We evaluate our retrieval-based generation model on 57 Julich-Brain areas selected to cover a diverse and representative set of cortical regions while balancing compute and annotation effort.
We use a structured LLM-based questionnaire to select the best-performing language backbone.
All image data are drawn from BigBrain~\cite{Amunts2013}, a high-resolution three-dimensional reconstruction of a single postmortem human brain that serves as a histological reference brain underlying the Julich-Brain atlas.
Because CytoNet operates on the Julich-Brain atlas and BigBrain is a histological reference within this same atlas framework, this choice provides a tightly coupled and well-grounded mapping between image patches, cytoarchitectonic labels, and literature-derived descriptions~\cite{Amunts2013,Amunts2020,schiffer2025cytonet}.
We additionally train an \emph{enriched} instruction-tuned model with patch-specific targets to evaluate whether the vision--language interface can recover local properties rather than only verbalize area-level information.

Our results show that weak, retrieve-and-enrich pairing can couple cytoarchitectonic vision foundation models to language, providing a practical route to natural-language integration in biomedical domains lacking curated image--text pairs.
We contribute:
1) A weakly supervised, \emph{retrieval-informed} vision--language model that couples CytoNet embeddings to natural-language cytoarchitectonic descriptions across 57 Julich-Brain areas.
2) An \emph{enriched} instruction-tuned model, built with the same base methodology, that additionally extracts and reasons about  image-specific cytoarchitectonic properties such as cortical layer thickness and cell density.
3) A literature distillation pipeline that extracts canonical statement pools per label and synthesizes text for training without curated image-level descriptions.

Existing alternatives to curated pairs include mining naturally occurring image–text pairs from literature figures, educational videos, or social media~\cite{ikezogwo2023quilt,huang2023visual}; bootstrapping captions from an existing captioner~\cite{li2022blip}; and text-only or unpaired training that exploits a joint contrastive image–text space~ \cite{nukrai2022text,feng2019unsupervised}. None applies directly here: there is no large-scale resource of cytoarchitectonic patches with associated text to mine, no in-domain captioner exists to bootstrap from, and CytoNet is trained on images alone, so there is no shared image–text space to decode.

\section{Method}

Our approach creates paired image--text data using a process of \emph{retrieve} and \emph{enrich}.
Labels are applied to both image and text and for each image, associated text is retrieved. This text is enriched by applying image specific data to it.

We aim to couple CytoNet~\cite{schiffer2025cytonet} to an LLM to generate descriptions of microscopic image patches.
We start from cell-body--stained histological human brain sections and extract image patches following the CytoNet pipeline.
To obtain dense supervision at scale, we apply CytoNet to assign each image patch a label for the corresponding Julich-Brain area~\cite{Amunts2020}, and to predict key anatomical features.
To obtain labels at scale, we apply CytoNet to assign each image patch labels for the corresponding Julich-Brain area~\cite{Amunts2020}.
In addition, we use CytoNet to prdict key anatomical features, including the thickness and cell density within the six cortical layers of the cortex - features that vary according to the structure and functionality of different areas of the brain.
In parallel, we construct area-specific textual supervision by mining neuroanatomical literature linked to these areas and extracting canonical statements about cytoarchitecture, which are combined into synthetic descriptions.

 \begin{figure}[t]
     \centering
     \includegraphics[width=0.9\linewidth]{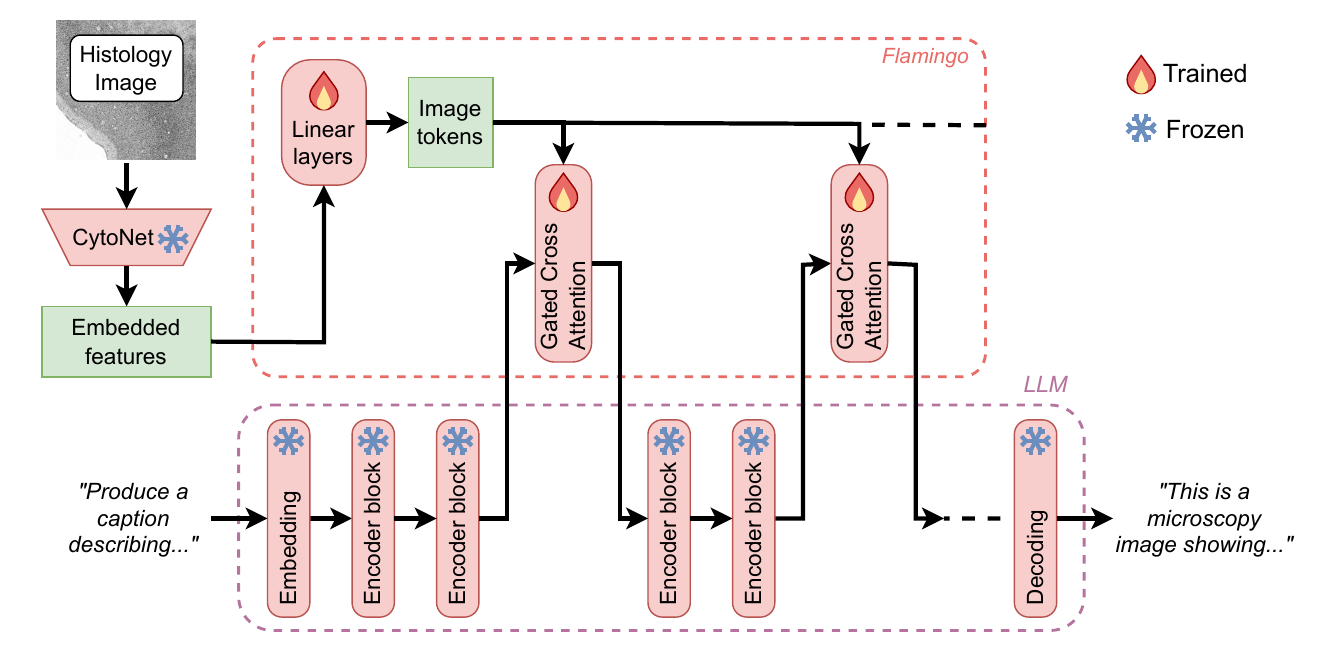}
     \caption{\textbf{Architecture overview.}
     Image-conditioned text generation is based on Flamingo~\cite{alayrac2022flamingo}.
     Condition images are embedded by the frozen vision model, CytoNet~\cite{schiffer2025cytonet}.
     Embeddings are  linearly projected into image tokens.
     A gated cross-attention block~\cite{alayrac2022flamingo} after every 4th encoder block of the large langauge model conditions text generation.
     }
     \label{fig:model_arch}
\end{figure}

We then train a Flamingo-style~\cite{alayrac2022flamingo} vision--language adapter that conditions a frozen large language model (Llama-3-8B~\cite{grattafiori2024llama3herdmodels} for the retrieval-based generation experiment, and Qwen3-4b~\cite{yang2025qwen3} for the instruct-tuned experiment)  on frozen CytoNet embeddings, using the labeled areas to weakly pair image patches with synthetic descriptions of canonical features of the area.
We use a Flamingo-style interface because it is a well-established way to condition a frozen language model on frozen visual features while training only a small cross-modal adapter. This matches our setting: the visual encoder already contains cytoarchitectonic structure, the paired data are synthetic and limited, and our compute budget does not support end-to-end LLM training.
A repository containing the workflows described and example histological image patches is available at: \url{https://jugit.fz-juelich.de/inm-1/bda/personal/msutton/cyto_in_words}.

\textbf{Image patch extraction from histology data.}
We use microscopic scans of cell-body--stained histological sections from BigBrain~\cite{Amunts2013}.
Images are processed in a patch-based fashion, with patches centered along the midline of the cortical sheet.
Following~\cite{schiffer2025cytonet}, we generate $\sim$539k image patches of size $2048\times2048$ pixels at $2~\si{\micro\metre}$/pixel ($4\times4~\si{\milli\metre}$ field of view). These were filtered to the 57 Julich-Brain areas following label assignment, and supplemented with image patches drawn from unknown areas in a ratio of 10:1.
This gave 192,000 image patches for training, 960 validation, and 3,000 for testing.

\textbf{Cytoarchitectonic labels with CytoNet.}
To obtain area labels for weak supervision, we apply a CytoNet-based area classifier~\cite{schiffer2025cytonet} to each extracted image patch and use its prediction as a label.
The classifier is trained to predict 157 Julich-Brain areas~\cite{Amunts2020}, enabling dense labeling across the cortical sheet without manual image-level annotation.
We restrict training to 57 target areas, and treat images predicted as any other area as \enquote{unknown}.

\begin{figure}[t]
    \centering
    \includegraphics[width=1\linewidth]{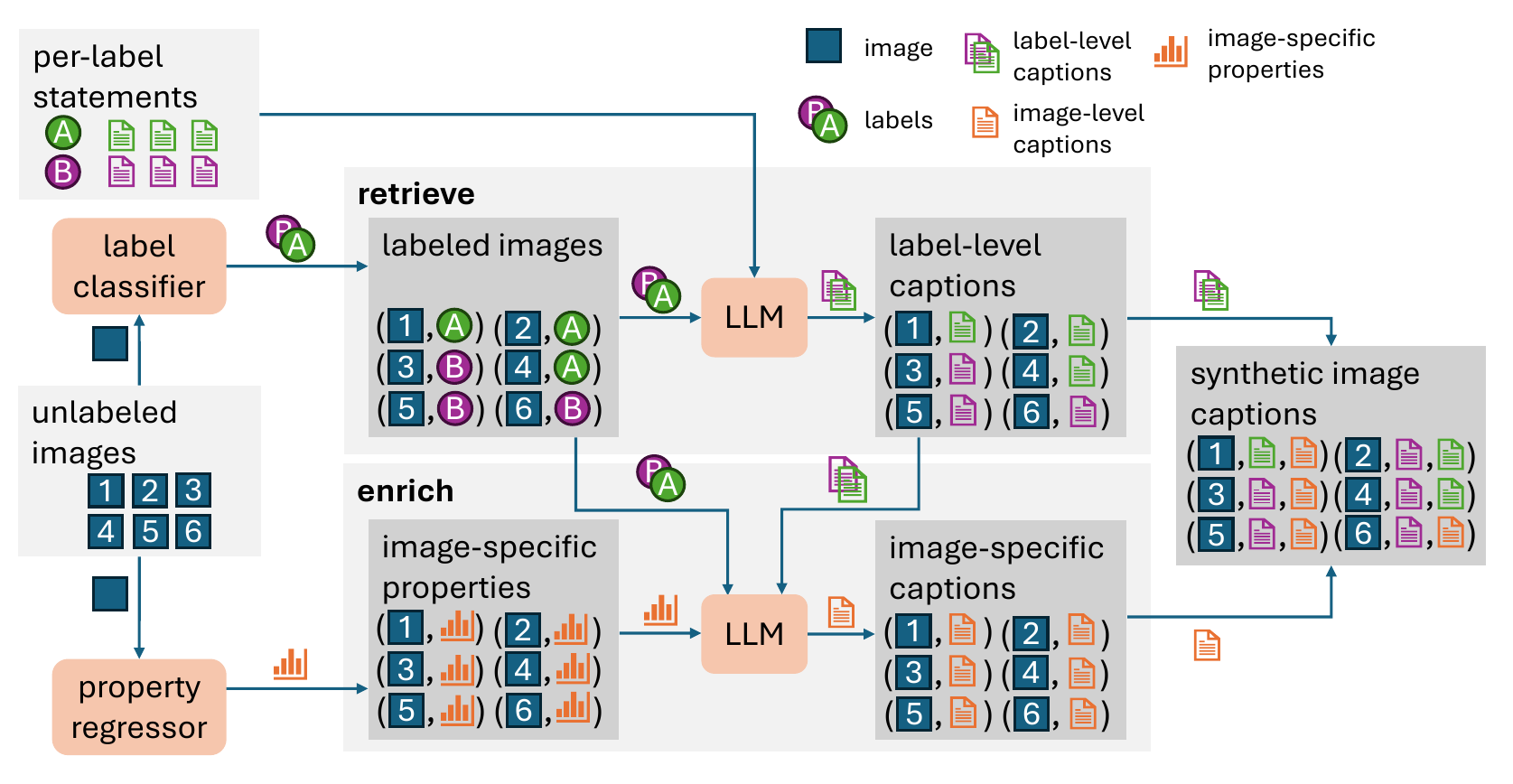}
    \caption{
    \textbf{Retrieve-and-enrich supervision.}
    Starting from unlabeled images (e.g., microscopic image patches), a label classifier assigns a label to each image patch.
    In the \emph{retrieve} stage, this label indexes literature-derived statement pools for the corresponding label, and an LLM converts sampled statements into label-level descriptions.
    In the \emph{enrich} stage, a property regressor provides image-specific measurements (e.g., cortical layer thickness and cell density), which are verbalized by an LLM as image-specific description.
    The final synthetic description combines label-level literature supervision with patch-specific properties.
    }    
\label{fig:scraping}
\end{figure}

To obtain patch-specific targets, we use a CytoNet-based property regressor for cortical layer thickness and cell density.
The regressor was trained on annotated histological image patches from the same brain \cite{patches_44,patches_4a,patches_SPL,patches_fg1,patches_fg3,patches_fg4,patches_fp1,patches_hOC1,patches_hOc2,patches_hOc3d,patches_hOc4d,patches_hOc5}.
The annotations include segmentations of the cortical layers and cell segmentation, allowing for the descriptive statistics of average thickness and cell density of each cortical layer to be calculated.
Additional patches in the same format, covering more brain areas from the BigBrain, were made available for this project.
In total, the patch-specific property experiment uses 902 image patches.

\textbf{Literature retrieval by cytoarchitectonic brain area.}
To obtain textual supervision for each cytoarchitectonic area, we construct an area-linked literature corpus (\cref{fig:scraping}).
The EBRAINS Knowledge Graph (KG) (\url{https://search.kg.ebrains.eu}) is a multimodal metadata store that brings together information from different fields of brain research, with its core being a graph database linking neuroscientific research across modalities.
We use the EBRAINS Knowledge Graph to automatically identify seed publications associated with Julich-Brain areas via links to probabilistic cytoarchitectonic maps of the Julich-Brain atlas and related reference works.
We then expand this set using citation search (Scopus) and filter candidate papers using area-specific keywords in titles/abstracts, excluding non-human and non-cytoarchitectonic studies.
Full texts available via PubMed or ScienceDirect are downloaded and converted to plain text for downstream processing, yielding 575 papers spanning the 57 areas.

\textbf{Area-specific statement extraction from literature.}
To derive area-specific textual supervision, we transform full-text papers into collections of short, stand-alone cytoarchitectonic statements that can serve as building blocks for descriptions of the canonical features of an area.
We convert the retrieved papers into plain text and split them into chunks to fit in memory.
For each chunk, we use Qwen3-Next~\cite{yang2025qwen3} to extract statements about the cytoarchitecture of the target area, explicitly avoiding paper-specific experimental details and requiring that each statement is interpretable without additional context.
The extracted statements are post-processed to strip formatting and present in a convenient, machine-readable format to be used for synthetic description generation.

\textbf{Synthetic description generation.}
To create synthetic image--text pairs, we create the descriptions from the statement pools for each of the 57 Julich-Brain areas.
For each image patch, a CytoNet-predicted Julich-Brain area label is used and statements associated with that area are randomly sampled.
We provide the statements together with the area name to Llama-3-8B-Instruct~\cite{grattafiori2024llama3herdmodels} to generate a concise description describing cytoarchitectonic properties of the depicted region.
This yields weak, area-conditional image--text pairs used to train the vision--language model.
For image patches sampled as an unknown label, a standard statement is provided, stating the image is from an unknown area.

\textbf{Synthetic image--text pair generation with enrichment.}
To create a dataset with user prompts and answers informed by supplied image patches, we synthesize question-answer pairs for each sample.
As before, a Julich-Brain area label is predicted for each image patch and matching statements are retrieved based on the label.
These are then enriched by adding the predicted average thickness and cell density of each cortical layer in the image patch.
This information is provided to Qwen3-32b~\cite{yang2025qwen3} through the system prompt, along with a user prompt containing a question about the architecture and/or brain area depicted in the image.
The question for each image patch is selected at random from a bank of 45 pre-written questions.
`Thinking' is enabled for the model, meaning that chain-of-thought reasoning is generated before the final answer.
This reasoning is also stored so that it can be distilled into the smaller trained model.
After training samples are generated, the system prompt is stripped and replaced with a prompt that does not contain information about the image patch, so that the model must derive this information directly from the image.
Because these samples involve longer, more complex queries and reasoning chains, which raise the memory and compute cost of training, we generated $10{,}000$ question--answer pairs for training, $100$ for validation, and $250$ for testing.

\textbf{Vision--language architecture.}
We adopt a Flamingo-style~\cite{alayrac2022flamingo} design that conditions a frozen language model on visual features via a lightweight trainable interface.
We use CytoNet-ViT-1M~\cite{schiffer2025cytonet} as a frozen vision encoder to embed each microscopy image patch into a cytoarchitecture-sensitive feature vector.
A small projection module (2 linear layers separated by GELU activation) maps this vector into 4 vision tokens, which we found to be sufficient and computationally more efficient than the original implementation~\cite{alayrac2022flamingo}.
For the enriched implementation, the projection module is given an additional regression head that predicts the z-score-normalized spatial coordinates of each image patch together with its cortical layer thicknesses and cell densities.
This head is used only during training, as an auxiliary objective that helps align the representation space.
It is not required at inference time.
We then use either Llama-3-8B-Instruct~\cite{grattafiori2024llama3herdmodels} or Qwen3-4b~\cite{yang2025qwen3} as a frozen generator and insert gated cross-attention blocks after every fourth transformer block, using the language hidden states as queries and the visual tokens as keys/values.
Only the projection module and gated cross-attention parameters are trained.

\textbf{Training objective and protocol.}
We train the vision--language method on weakly paired prompt--response data, using CytoNet embeddings as visual inputs and synthetic descriptions or generated question-answers as targets.
Given an instruction prompt (e.g., \enquote{Provide a caption for this microscopy image of the human brain, describing its cytoarchitecture.}), the models are optimized to predict the target caption tokens. 
We use cross-entropy loss over generated tokens and keep both CytoNet and the language model frozen during training. In addition, for the enriched training, an MSE loss is applied to the secondary head's predictions, and is added to the cross-entropy loss.
Training of the retrieval-based generation model is run for 6 epochs with learning rate $10^{-3}$ and batch size 320 on 32 Nvidia A100 GPUs (8 compute nodes) using fully-sharded data-parallel (FSDP) training.
Training of the instruct-tuned model is run for 6 epochs with learning rate $10^{-4}$ and batch size 4 on 2 Nvidia H100 GPUs using fully-sharded data-parallel (FSDP) training.
Training was performed on JURECA-DC~\cite{Krause2018} at Jülich Supercomputing Centre.

\textbf{Cytoarchitectonic question/answer benchmark generation.}
A practical question for cytoarchitectonic vision--language modelling is which publicly available, open-weight LLMs are suitable backbones for adaptation, and we address this systematically:
To this end, we generate a multiple-choice question/answer (QA) benchmark from the retrieved neuroanatomical literature corpus to probe cytoarchitectonic knowledge and reasoning.
We extract text from PDFs, split documents into context-sized chunks, and prompt an instruction-tuned LLM to produce questions targeting cytoarchitectonic attributes (e.g., density, lamination) while avoiding paper-specific experimental details. Qwen3-Next was the chosen for this task over a larger model for reasons of the required compute. 
We post-process outputs by randomizing answer option order and filtering position-dependent artifacts, yielding 1{,}735 questions used to compare candidate LLM backbones and guide model selection. 

\section{Results}

\begin{table}[t]
    \centering
    \caption{
        \textbf{Performance of LLMs on the created QA dataset for cytoarchiteture.}
        Scores are the share of correctly answered multiple choice questions, out of a set of 1{,}735. For comparison, the first author answered a random subset of 50 questions.
        }
    \begin{tabular}{lccc}
        Model & \# Parameters & \makecell{Medical\\fine-tuning?}& Score \\
         \hline
        First author performance & n.a. & n.a. & 32.0\%\\
        BioGPT \cite{10.1093/bib/bbac409}& 0.35b & \cmark & 24.1\%\\
        BioGPT-Large \cite{10.1093/bib/bbac409}& 1.5b & \cmark & 26.3\%\\
        Qwen3 4b~\cite{yang2025qwen3}& 4b & \xmark & 54.1\% \\
        PMC Llama 7b \cite{wu2024pmc}& 7b & \cmark & 31.8\% \\
        Llama-3-8b \cite{grattafiori2024llama3herdmodels}& 8b & \xmark & 61.7\%\\
        Ministral 8b \cite{liu2026ministral}& 8b & \xmark & 55.2\% \\
        Llama-3-8b Ultramedical \cite{10.5555/3737916.3738735}& 8b & \cmark & 44.8\%\\
        Phi-4 \cite{abdin2024phi}& 14b & \xmark & 66.1\%\\
        Llama-3-70b \cite{grattafiori2024llama3herdmodels}& 70b & \xmark & 67.3\%\\
        Llama-3-70b Ultramedical \cite{10.5555/3737916.3738735}& 70b & \cmark & 54.2\%\\
        Qwen3-Next \cite{yang2025qwen3}& 80b & \xmark & 73.5\%\\
    \end{tabular}
    \label{tab:CytoQAScores}
\end{table}

\subsection{LLM backbone selection via cytoarchitectonic QA}
We first asked which publicly available, open-weight LLMs provide a suitable backbone for cytoarchitectonic vision--language modelling.
Using the cytoarchitectonic QA benchmark (1{,}735 multiple-choice questions), we evaluated a range of general-purpose and biomedical/medical-tuned models (Tab.~\ref{tab:CytoQAScores}).
General-purpose instruction-tuned models performed best on this domain-specific benchmark, with larger models performing better.
Biomedical/medical fine-tuned variants did not consistently improve performance over their base models.
Based on this screening, we selected Llama-3-8B-Instruct as the language backbone for the retrieval model, balancing between performance and computational requirements. Qwen3-4b~\cite{yang2025qwen3} was selected for the enriched model, due to its smaller size given larger context requirements during training and reasoning capabilities, despite lower performance on the benchmark.

\begin{figure}[t]
    \centering
    \includegraphics[width=0.95\linewidth]{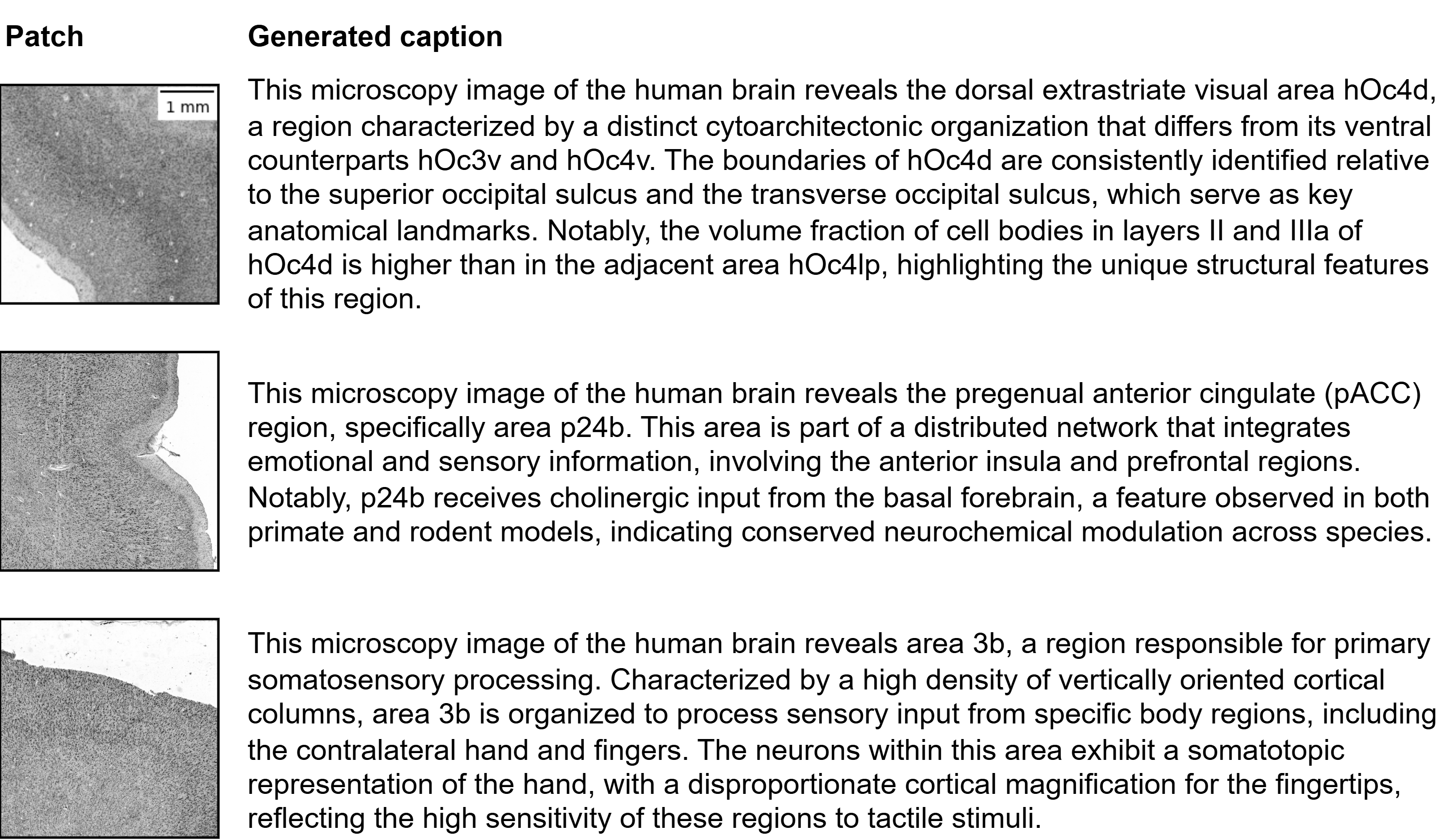}
    \caption{
    \textbf{Examples for generated descriptions of given microscopic image patches.}
    The text indicates the respective brain areas, as well as descriptions of area-specific characteristics.
    }
    \label{fig:example_captions}
\end{figure}

\subsection{Quality of generated descriptions}
Quantitatively assessing cytoarchitectonic caption quality is challenging, as standard captioning metrics do not test neuroanatomical validity, and no established image-level caption benchmarks exist for this domain.
For the caption-generation model, which is trained on retrieved area-level text only, we designed two complementary tests.
The \emph{label consistency test} quantifies whether the area named in the generated caption matches the ground-truth area.
The \emph{label-masked description discriminability test} quantifies whether the caption's description alone, once every area name is removed, still identifies the correct area.
A third evaluation, reported separately below, instead probes the \emph{enriched} model and asks whether it can read out patch-specific properties rather than only canonical area descriptions.
Confidence intervals were estimated by 10k bootstrap iterations.

\textbf{Label consistency.}
Captions follow a standardized format in which the first sentence specifies a Julich-Brain area (or \enquote{unknown}).
We automatically extract this label and compare it to the reference label predicted by CytoNet.
On image patches from target areas, extracted labels match the reference label in 90.6\% of cases (95\% confidence interval (ci): 88.9\%--91.2\%); for out-of-scope images, \enquote{unknown} is produced with 91.4\% accuracy (95\% ci: 8
8.1\%--94.4\%); the overall F1=0.82.

\textbf{Description discriminability with label masking.}
For each generated caption, we automatically mask out all mentions of brain areas, leaving only descriptions of cytoarchitectonic properties.
This redacted text is presented to Qwen3-Next in an 8-way multiple-choice setting.
The eight options consist of the area label extracted in the label
consistency test and seven distractor areas sampled uniformly at random from the 57 target areas (excluding unknown).
To reduce reliance on the judge model’s prior cytoarchitectonic knowledge, each candidate area is accompanied by five literature-derived statements for that area from our statement pool.
Qwen3-Next is asked which of the eight areas the redacted description refers to.
We define the correct answer as the area predicted in the first test (rather than the underlying reference label) to decouple label correctness from caption quality.
Qwen3-Next selects the correct area in 68.6\% of cases (95\% ci: 66.8\%--70.4\%), indicating that the generated captions are identifiable in a majority of cases, and significantly above chance (12.5\%).

\begin{figure}[t]
    \centering
    \includegraphics[width=\linewidth]{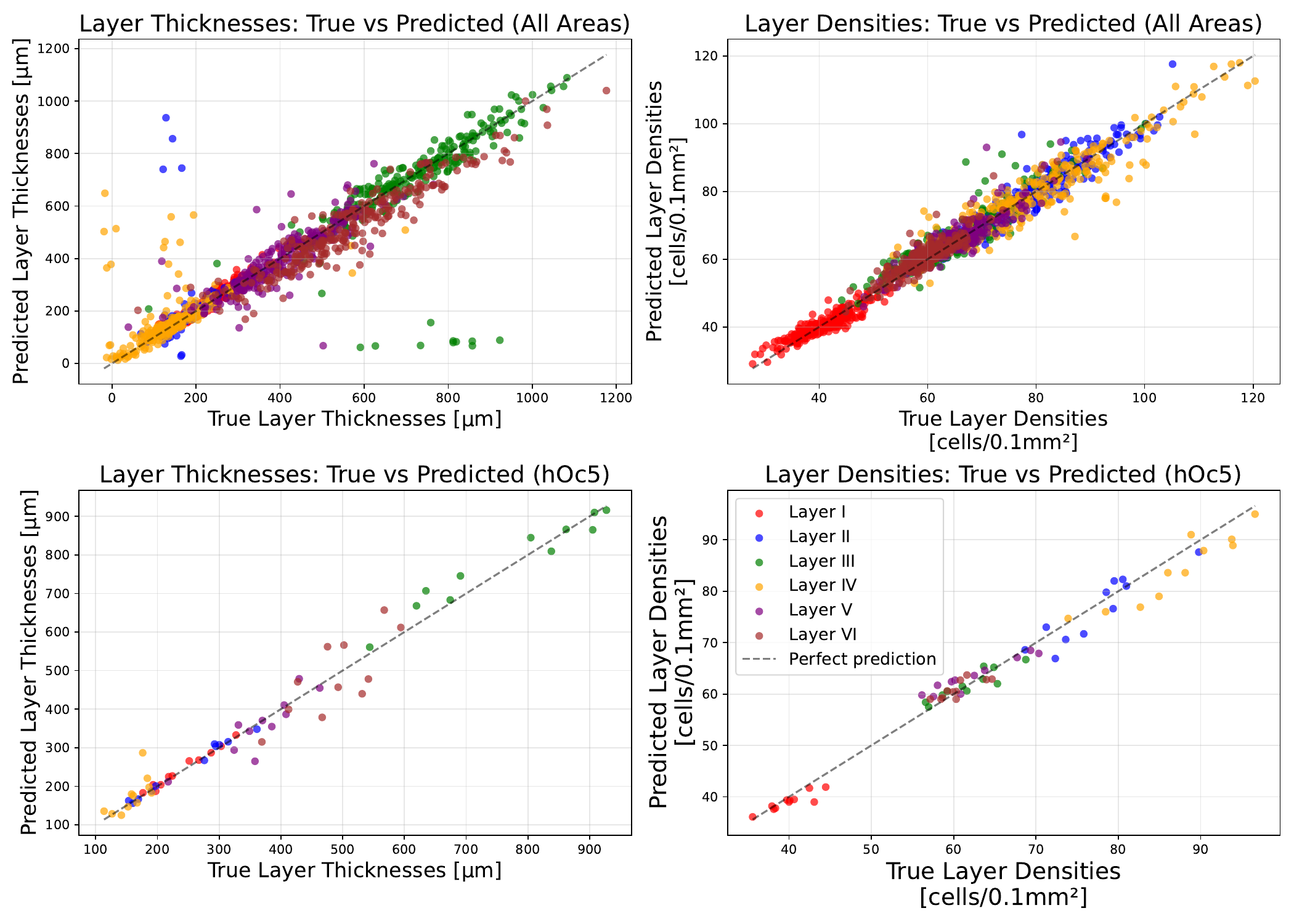}
    \caption{
    \textbf{Predicted cortical layer thicknesses and densities against ground truths for all considered areas (top) and area hOc5 (bottom).}
    The model can accurately predict cortical layer thicknesses and cell densities, with  individual outliers in the thicknesses of layers II, III and IV.
    The results for area hOc5 demonstrate that the model is able to accurately capture intra-area variations.
    }
    \label{fig:layer_thickness_density}
\end{figure}

\textbf{Recovering patch-specific properties with enrichment.}
%The previous tests evaluate whether generated captions are consistent with area-level supervision.
To test whether the vision--language interface can recover information specific to an individual image patch, we train a second, instruction-tuned model whose targets additionally include the cortical layer thicknesses and cell densities measured in that patch.
For each image in the test dataset, the model is prompted to identify the brain area and to state the thickness and density of each cortical layer visible.

The area-identification performance of this model is lower than that of the caption-generation model, which we attribute to its considerably smaller training dataset: it identifies the correct area in 33.3\% of cases.
Following~\cite{schiffer2025cytonet}, we characterize the errors in more detail using the area adjacency graph of the Julich-Brain atlas (v3.2,~\cite{Amunts2020}), in which two areas are one hop apart if they physically border each other.
Of the incorrect predictions, 55.4\% fall within one hop of the true area and 75.3\% within two hops.
This indicates that the model generally identifies approximately the correct brain region while struggling to recover the more fine-grained local subdivisions.
Based on the results of the caption-generation model, we expect that substantially longer training on a larger dataset would help close this performance gap.
The model recovers the average thickness and cell density of the cortical layers accurately in most cases, showing it is able to make image-specific determinations (\cref{fig:layer_thickness_density}).
Substantial outliers remain, however, particularly in the predicted thickness of layers II, III and IV.
The plots restricted to area hOc5 confirm that these values are not derived purely from the area label: the model also reproduces the variation \emph{within} a single area.
This is clearest for the thickness of layer III and the density of layer IV.

\subsection{Comparison to a template-based baseline}
\label{sec:template}

\begin{figure}[h!]
	\centering
	\includegraphics[width=0.9\linewidth]{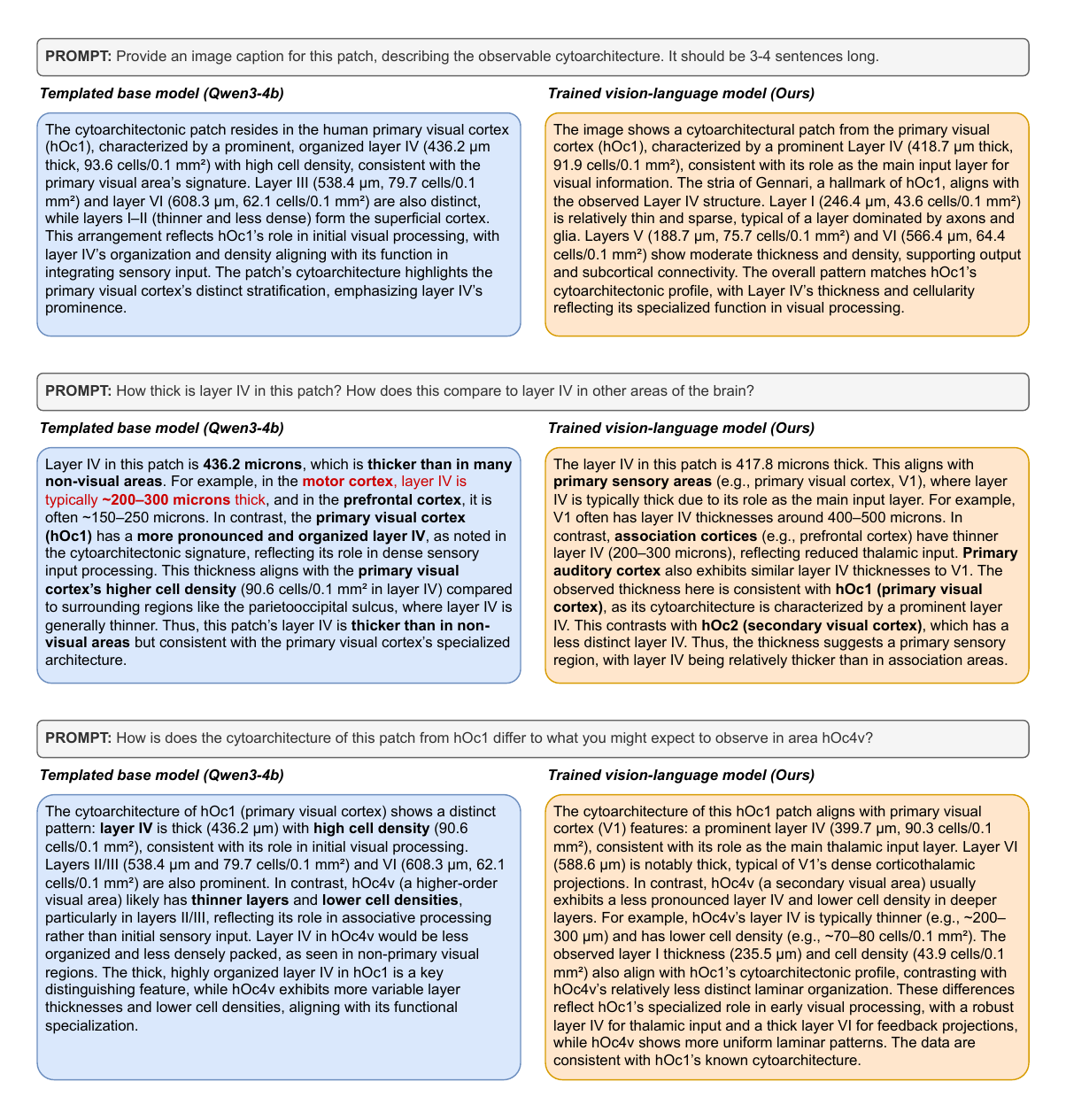}
	\caption{
		\textbf{Comparison of outputs to the same prompt from the base language model (Qwen3-4b) and from our trained vision--language model.}
		Qwen3-4b~\cite{yang2025qwen3} was provided with a template of information replicating what was supplied to the model during generation of the synthetic training data. Our model was shown only the associated image patch and given no further information.
		On tasks involving reasoning about other areas of the brain, our trained model shows more certainty and precision, demonstrating a broader understanding of cytoarchitecture.
		The base model made a substantial error when discussing the motor cortex, highlighted in red.
	}
	\label{fig:template_comp}
\end{figure}

A remaining baseline question is how much of the observed behavior can be reproduced by a simple templating approach, in which an untrained language model receives a text template derived from the predicted labels and properties and the associated statement pool.
A rigorous comparison is non-trivial as no established benchmark exists for image-level cytoarchitectonic captioning.
As an illustrative comparison, Fig.~\ref{fig:template_comp} prompts the base Qwen3-4B model with the same type of template information used to synthesize responses for a primary visual cortex patch, while the trained vision--language model receives only the image and the prompt.
Conversation history is cleared between prompts.
Both models can produce plausible captions, but the trained model gives more specific and distinct answers when asked to reason about relationships to other areas.

\section{Discussion \& conclusion}

This work targets domains where expert annotation is scarce but textual knowledge is abundant, and applies a retrieve-and-enrich weak supervision scheme to connect cytoarchitectonic vision foundation models to language.
The presented method learns a vision--language coupling that recovers area-level and image-specific cortical layer thicknesses and densities and generates plausible cytoarchitectonic descriptions despite the absence of curated image--text pairs.
This is a step towards enabling natural language interaction with human brain atlases, providing a valuable tool for researchers to accelerate their work and paving the way for agentic research workflows.
A central practical result is that selecting a suitable language backbone for cytoarchitectonic tasks benefits from systematic screening: our automatically constructed cytoarchitectonic QA benchmark reveals substantial variation in domain competence across open-weight models and provides actionable guidance for backbone selection.

Methodologically, the workflow is scalable, but makes end-to-end validation difficult because the training texts are synthetic rather than expert-written.
We therefore evaluate complementary behaviors: area consistency, description discriminability after label masking, and recovery of patch-specific properties.
This design has three limitations.
First, discrete area labels do not fully capture gradual transitions or mixed tissue near areal borders.
Second, labels are derived from CytoNet predictions rather than manual image-level annotations, providing whole-cortex coverage but potentially introducing label noise.
Third, experiments are conducted on BigBrain~\cite{Amunts2013}; extending to additional brains is necessary to assess cross-subject transfer and robustness to inter-individual variability.

Future work should compare Flamingo-style adaptation with other established interfaces between frozen visual encoders and language models, such as LLaVA~\cite{liu2023visual} or BLIP-2~\cite{li2023blip}.
The evaluation could also be extended with expert qualitative review or blinded preference comparisons against suitable baselines, because expert judgment remains the most direct way to assess cytoarchitectonic correctness of free-form generated text.
Extending annotations to all Julich-Brain areas would broaden coverage.
In parallel, incorporating literature resources (e.g., historical atlases~\cite{VonEconomo1925}) and targeted expert annotation of  image subsets could provide high-quality anchors.
Diversification of the studied brains, for factors such as a disease or age of the individual, would further expand the scope of the model. However, there are inherent limitations in postmortem brain histology due to the limited possible sample sizes given the practical constraints of tissue acquisition and processing, so this will likely to continue to be difficult to address.
The introduction of \enquote{tools} that the model can access in an agentic fashion for some predictions could aid in precision and allow the model to focus on reasoning with information rather than the pure prediction task.

Overall, we present a practical recipe for biomedical imaging domains where images and text exist at scale but curated image--text pairs are unavailable: combine dense retrieve-and-enrich alignment with literature distillation and lightweight cross-modal adaptation.
For example, a labeled image dataset of CT scans depicting a range of liver conditions could be combined with literature on each ailment to provide similar captioning.
Using open-weight models improves reproducibility and version control while still enabling strong performance in this specialized setting.
We expect these ideas to generalize beyond cytoarchitecture to other microscopy and histopathology domains where expert concepts are well documented but difficult to attach to individual images at scale.

% offloading some stuff as tools

% Note: if you want to use up all of the allowed space for the paper,
%       the bibliography will start on top of page 13. Furthermore,
%       from page 13 onwards, there will be *only* bibliography, no more
%       figures/tables.
\bibliographystyle{splncs04}
\bibliography{093-main}

\end{document}